\documentclass[letterpaper, 10 pt, conference]{ieeeconf}  

\IEEEoverridecommandlockouts                              

\usepackage{amsmath}
\usepackage{graphicx}
\usepackage{float} 
\usepackage{cleveref}

\title{\LARGE \bf
Adaptive Prior Scene-Object SLAM for Dynamic Environments
}

\author{Haolan Zhang, Thanh Nguyen Canh, Chenghao Li, and Nak Young Chong
\thanks{*This work was supported by the Asian Office of Aerospace Research and Development under Grant/Cooperative Agreement Award No. FA2386-22-1-4042, and  JST SPRING, Japan Grant Number JPMJSP2102.}
\thanks{The authors are with the School of Information Science, Japan Advanced Institute of Science and Technology,
        Ishikawa 923-1292, Japan. {\tt \{s2420423,thanhnc,chenghao.li,nakyoung\}@jaist.ac.jp}}%
}

\begin{document}

\maketitle
\thispagestyle{empty}
\pagestyle{empty}

\begin{abstract}

Visual Simultaneous Localization and Mapping (SLAM) plays a vital role in real-time localization for autonomous systems. However, traditional SLAM methods, which assume a static environment, often suffer from significant localization drift in dynamic scenarios. While recent advancements have improved SLAM performance in such environments, these systems still struggle with localization drift, particularly due to abrupt viewpoint changes and poorly characterized moving objects. In this paper, we propose a novel scene-object-based reliability assessment framework that comprehensively evaluates SLAM stability through both current frame quality metrics and scene changes relative to reliable reference frames. Furthermore, to tackle the lack of error correction mechanisms in existing systems when pose estimation becomes unreliable, we employ a pose refinement strategy that leverages information from reliable frames to optimize camera pose estimation, effectively mitigating the adverse effects of dynamic interference. Extensive experiments on the TUM RGB-D datasets demonstrate that our approach achieves substantial improvements in localization accuracy and system robustness under challenging dynamic scenarios.

\end{abstract}

\section{INTRODUCTION}

Simultaneous Localization and Mapping (SLAM) enables autonomous systems to navigate unknown environments, with visual SLAM gaining prominence due to the cost-effectiveness of cameras. Traditionally, visual SLAM has evolved into two 
methodologies: feature-based methods~\cite{c1,c2,c3}, which extract and match distinctive keypoints, and direct methods~\cite{c4, c5}, which operate on pixel intensities directly without intermediate feature extraction. While these approaches are effective in static environments, they face significant challenges in dynamic scenarios due to moving objects and abrupt viewpoint changes, leading to localization drift.

Dynamic scenarios lead to incorrect feature matching and localization drift. Early solutions focused on geometric approaches, such as RANSAC-based filtering (Sun \emph{et al}.~\cite{c6}), probabilistic confidence scoring (Li \emph{et al}.~\cite{c7}), and correlation-based graph segmentation (Dai \emph{et al}.~\cite{c8}). More recent motion-based methods emerged, including PFD-SLAM~\cite{c9} with optical flow and particle filtering, as well as StaticFusion~\cite{c10} and Joint-VO-SF~\cite{c11}, which applied K-means clustering for static probability estimation. While effective in mildly dynamic scenes, these methods struggle to handle large-scale dynamic changes.

To overcome these limitations, researchers have integrated deep learning techniques. DS-SLAM~\cite{c12} leverages SegNet for semantic filtering, while DynaSLAM~\cite{c13} combines Mask R-CNN with geometric verification. Blitz-SLAM~\cite{c14} adopts a two-stage approach, first parsing scenes with deep learning, followed by geometric validation. CFP-SLAM~\cite{c15} employs hierarchical processing based on object detection and motion classification, and SG-SLAM~\cite{c16} integrates semantic understanding with geometric constraints in a graph-based framework.

Building upon these approaches, recent works have explored object-centric and scene-centric strategies. Liu \emph{et al}.~\cite{c17} introduces an object-centric method that evaluates quality based on uncertainty, observation quality, and prior information, implementing dual coupling where high-quality objects contribute to camera pose estimation while low-quality objects are only tracked afterward.

However, object-centric approaches struggle with fixed quality thresholds that fail to adapt to scene variations, frame-by-frame evaluation that ignores temporal consistency, and the absence of error correction. Long \emph{et al}.~\cite{c18} introduced a scene-centric method leveraging prior motion information to enhance temporal consistency, but it struggles with sudden motion changes, risks skipping critical frames due to reliance on previous frames, and may misclassify scenes with low-motion as static.

To overcome the limitations of both object-centric and scene-centric approaches, we propose an Adaptive Prior Scene-Object SLAM framework for dynamic environments based on ORB-SLAM3~\cite{c19}. The key contributions are:
\begin{itemize} \setlength{\itemsep}{0.1em}
\item A scene-object quality assessment mechanism that integrates frame-based metrics with dynamic changes evaluation for reliable scene assessment.
\item An adaptive benchmark update strategy that continuously refines reference criteria based on scene quality.
\item A direct fusion method to correct pose estimation errors in problematic frames, enhancing robustness.
\item Extensive experimental validation on the TUM RGB-D dataset, demonstrating significant localization and robustness improvements.
\end{itemize}

The remainder of this paper is organized as follows: Section \uppercase\expandafter{\romannumeral2} presents our proposed framework, including the scene-object quality assessment mechanism and pose refinement strategy. Section \uppercase\expandafter{\romannumeral3} presents experimental results and comparative analysis. Finally, Section \uppercase\expandafter{\romannumeral4} concludes the paper with discussions and future directions.

\section{METHODOLOGY}

\begin{figure*}[t]
  \centering
  \setlength{\fboxsep}{0pt} 
  \setlength{\fboxrule}{0pt} 
  \fbox{\includegraphics[width=0.95\linewidth]{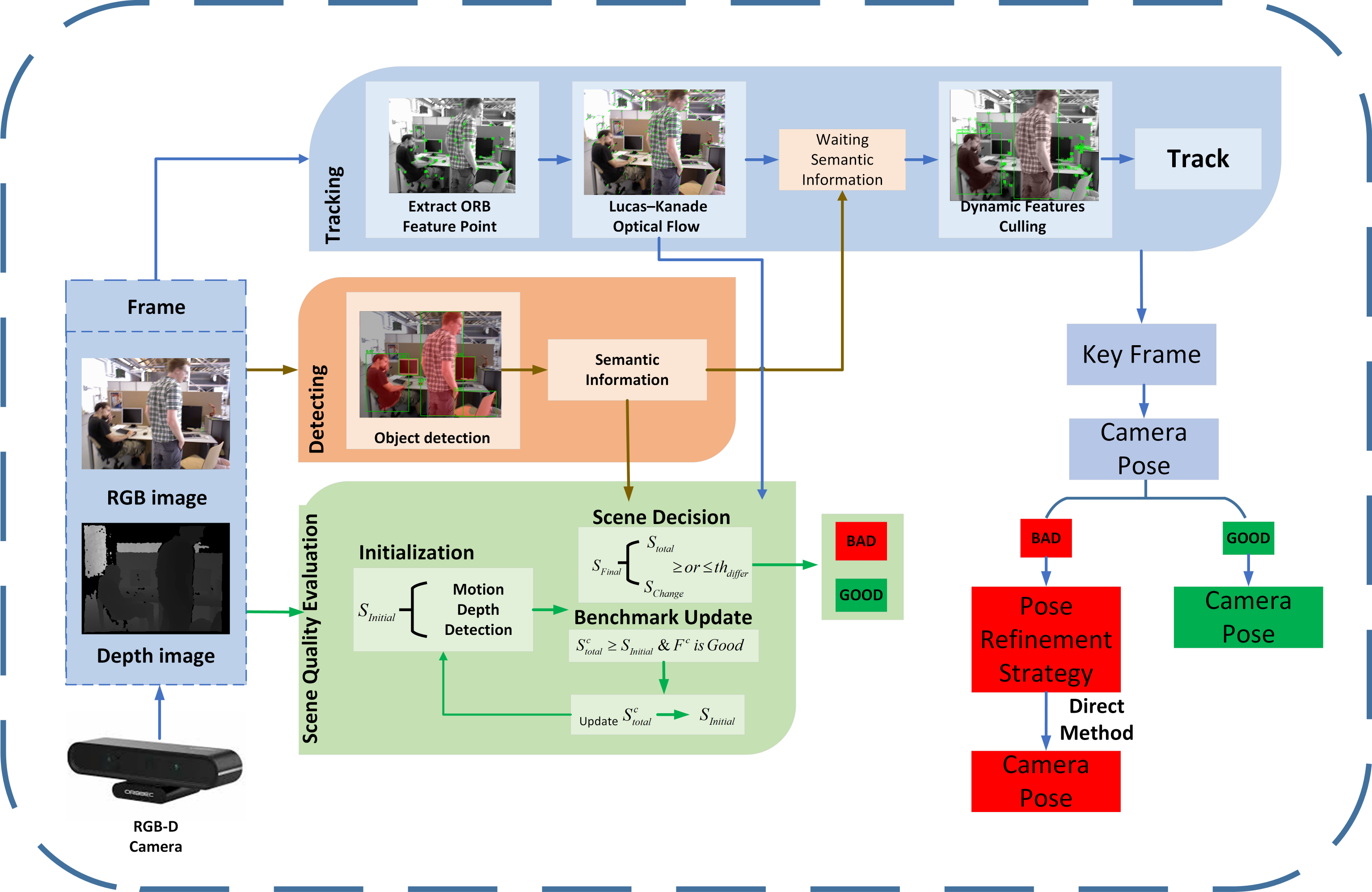}}
  \caption{\textbf{Overview of our proposed adaptive prior scene-object SLAM framework:} The framework is composed of three main units: Dynamic Observation module, Scene-Object Quality Assessment module, and Pose Refinement module.}
  \label{fig:framework}
\end{figure*}

Our proposed pipeline (Fig.~\ref{fig:framework}) processes RGB-D images through an adaptive framework for robust localization in dynamic environments. It combines feature extraction, semantic segmentation, and Lucas-Kanade optical flow to identify dynamic objects. Our approach consists of two key components: a scene-object quality assessment mechanism (Section \uppercase\expandafter{\romannumeral2}.A) and a pose refinement strategy (Section \uppercase\expandafter{\romannumeral2}.B). The quality assessment establishes baseline criteria, evaluates frames against benchmarks, and dynamically updates benchmarks as environmental conditions change. When frames are unreliable, the refinement strategy uses direct methods to correct pose estimation, maintaining robustness in dynamic scenarios.

\subsection{Scene-Object Quality Assessment Mechanism}
\subsubsection{Initialization}
The Scene-Object Quality Assessment mechanism begins with an initialization phase to establish reliable baseline criteria. Our system quantifies frame quality using four metrics: \textbf{Object Confidence Score} ($S_{conf}$): measures detection confidence across all objects, \textbf{Spatial Distribution Score} ($S_{spatial}$): assesses object size and position within the frame, \textbf{Feature Quality Score} ($S_{feature}$: evaluates feature response strength and distribution uniformity, and \textbf{Depth Quality Score} ($S_{depth}$): analyzes depth coverage, consistency and smoothness. The overall initial quality is calculated as:
\begin{equation}
S_{Initial} = f(S_{conf}, S_{spatial}, S_{feature}, S_{depth})
\label{eq:quality_score_initial}
\end{equation}
The function $f(\cdot)$ is a weighted combination:
\begin{equation}
f(S_{conf}, S_{spatial}, S_{feature}, S_{depth}) = \sum_{i} w_i S_i, \quad \sum_{i} w_i = 1
\label{eq:function_f}
\end{equation}
where $w_i$ is the weight for each component $S_i$.

During initialization, the system selects the frame with the highest quality score ($S_{Initial}$) as the reference, provided:
\begin{equation}
S_{Initial} \geq th_s \text{ and } N_{Initial} \leq th_f
\label{eq:Initial}
\end{equation}
where $th_s$ is the minimum quality threshold, $th_f$ is the maximum frame count, and $N_{Initial}$ is the reference frame number.

If no suitable frames are found, the system adaptively adjusts thresholds:
\begin{equation}
th_f = \min(th_{fmax}, th_f \cdot e^{\max(0, \frac{n}{th_f} - 1)})
\label{eq:Adaptive_Frame}
\end{equation}
\begin{equation}
th_{s} = \max(th_{smin}, \beta\cdot th_s \cdot log(\frac{th_f}{n}))
\label{eq:Adaptive_Quality}
\end{equation}
where $n$ is the current frame number and $\beta$ is a decay control coefficient that moderates the rate.

The initialization phase establishes a baseline for scene evaluation by analyzing object confidence, spatial distribution, feature quality, and depth information across initial frames. The highest-scoring frame becomes our reference benchmark, with adaptive initialization thresholds that adjust to maintain relevance as environmental conditions change.

\subsubsection{Scene Decision}
Following initialization, our system evaluates each incoming frame to determine scene reliability. The final scene quality score combines the current frame quality assessment ($S_{total}$)  and the change evaluation  ($S_{change}$) relative to the reference frame:
\begin{equation}
S_{final}=W_{base}S_{total}+W_{change}S_{change}
\label{eq:Final Score}
\end{equation}
where $S_{total}$ follows the same evaluation method as $S_{Initial}$ described in the initialization phase (Section \uppercase\expandafter{\romannumeral2}.A.1). $W_{base}$ and $W_{change}$ will be defined in Eq.~\ref{eq:W_base} and Eq.~\ref{eq:W_change}. The $S_{change}$ component quantifies the deviation from the reference frame:
\begin{equation}
S_{change}=f(S_{mc}, S_{dc}, S_{dec})
\label{eq:Change Score}
\end{equation}
where $f(\cdot)$ follows the same weighted combination function as defined in Eq.~\ref{eq:function_f}, applied to the change score components. This change score integrates three key residual measurements: \textbf{Motion Residual} ($S_{mc}$) evaluates the average motion deviation between frames, considering both point-level motion magnitude and grid-level motion across valid grids. \textbf{Depth Residual} ($S_{dc}$) captures changes in depth information by evaluating average depth change and the standard deviation of depth changes. \textbf{Detection Residual} ($S_{dec}$) measures changes in object detection by quantifying both object count differences and the average IoU (Intersection over Union) change between corresponding objects.

To enhance robustness in challenging scenarios, we introduce the $Dynamic\_ratio$ metric, representing the proportion of dynamic objects in the scene:
\begin{equation}
W_{base} = max(0, W_{base}-a\cdot Dynamic\_ratio)
\label{eq:W_base}
\end{equation}
\begin{equation}
W_{change} = 1-W_{base}
\label{eq:W_change}
\end{equation}
where $a$ is a robust factor. Our system identifies and applies specialized processing for three distinct scenarios: \textbf{Highly Static Scenes} ($Dynamic\_ratio \leq th_{static}$ and $S_{feature} \geq S_{static}$), where we boost the final score. \textbf{Highly Dynamic Scenes} ($Dynamic\_ratio\geq th_{dynamic}$), prioritizing change evaluation. And \textbf{High Confidence Detections} ($S_{conf} \geq th_{obj}$), amplifying scores for scenes with reliable object detections. Here, $th_{static}$ is the upper threshold for static scene classification, $th_{dynamic}$ is the lower threshold for dynamic scene classification, $S_{static}$ is the minimum feature quality required for static scenes and $th_{obj}$ is the confidence threshold for reliable object detection.

While evaluating incoming frames, our system simultaneously performs dynamic benchmark updates:
\begin{equation}
S^i_{total} > S_{Initial} \text{ \& } F^i \text{ is good scene} \xrightarrow{\text{update}} S_{Initial} = S^i_{total} 
\label{eq:Benchmark Update}
\end{equation}
where $F^i$ is the $i$\_th frame and $S^i_{total}$ is the frame quality score of the frame $F^i$. When $S^i_{total}$ is greater than the initial quality benchmark $S_{Initial}$ and the frame $F^i$ is good scene, then the system updates the initial quality benchmark to match the quality score of frame $F^i$.

After computing the final score $S_{final}$, our system determines scene reliability:
\begin{equation}
\text{Scene-Object Quality} = \begin{cases}
GOOD & \text{if } S_{final} \geq th_{differ} \\
BAD & \text{otherwise}
\end{cases} 
\label{eq:Final Decision}
\end{equation}
where $th_{differ}$ is the threshold value that determines whether a scene has sufficient quality for conventional tracking methods. This binary classification triggers different processing strategies: for reliable frames, conventional feature-based tracking continues, while problematic frames activate our pose refinement mechanism described in Section {\uppercase\expandafter{\romannumeral2}.B}.

\subsection{Pose Refinement Strategy}

When a frame is classified as problematic based on our quality assessment, we employ a direct method to refine the camera pose estimate. Unlike feature-based methods that rely on sparse correspondences, our approach utilizes dense information from both intensity and depth images, making it more robust in challenging scenarios.

The pose refinement process begins by selecting appropriate reference frames:
\begin{equation}
{\left\| {\left\| {D_i - {D_r}} \right\|_2 + \left\| {M_i - {M_r}} \right\|} \right\|_2} \le t{h_{keyframes}}
\label{eq:Reference Frame}
\end{equation}
where $D_i$ and $M_i$ represent the depth and motion characteristics of the current frame, while $D_r$ and $M_r$ correspond to previously identified good frames. The threshold $th_{keyframes}$ determines the maximum allowable combined difference for selecting a reliable reference frame.

Following previous RGB-D SLAM approaches \cite{c10,c18}, our direct method minimizes the intensity and depth residuals between image pairs. Specifically, for a pixel $p$ in frame A, the intensity and depth residuals with respect to frame B are defined as:
\begin{equation}
r_I^p = I_B(W(x_A^p, T(\xi), D_A)) - I_A(x_A^p)
\label{eq:Intensity residual}
\end{equation}
\begin{equation}
r_D^p = D_B(W(x_A^p, T(\xi), D_A)) - |T(\xi)\pi^{-1}(x_A^p, D_A(x_A^p))|_z 
\label{eq:Depth residual}
\end{equation}
where $I_A$ and $I_B$ are RGB images, $D_A$ and $D_B$ are depth images, $W$ is the warping function that projects pixels from frame A to frame B based on the transformation $T(\xi)$ and $\pi^{-1}$ represents the inverse projection from 2D to 3D space, $x_A^p$ is the 2D coordinate of pixel $p$ in frame A, and $|\cdot|_z$ denotes the z-component of the transformed 3D point.

The combined cost function incorporates both residuals:
\begin{equation}
D = C(\alpha_I w_I^p r_I^p(\xi)) + C(w_D^p r_D^p(\xi))
\label{eq:Combined cost function}
\end{equation}
where $w_I^p$ and $w_D^p$ are pixel-specific weights, $a_I$ is a scale parameter to weight photometric residuals so that they are comparable to depth residuals. $C(\cdot)$ is a robust cost function.

We adapt the optimization approach from \cite{c18} with modifications to handle dynamic environments more effectively:
\begin{equation}
R(\zeta, \gamma, k) = \sum_{p=1,k=1}^{N,K} \{B_k * \gamma_i(p)[D]\}
\label{eq:direct optimization approach}
\end{equation}
where $\gamma_i(p)$ represents pixel-specific weights that assign lower values to foreground (Potential Dynamic objects) pixels and higher values to background (Static) pixels. $B_k=e^{-\lambda(t_k-t_c)}$ is a time-decay factor that gives higher priority to more recent reference frames with $k$ is being the index of reference frames and $K$ is the number of reference frames. $[D]$ refers to the result from Eq.~\ref{eq:Combined cost function}.

The final camera pose is computed as a weighted combination:
\begin{equation}
T_{Final}=w_{diff}\cdot T_{feature} + (1-w_{diff})\cdot T_{direct}
\label{Optimized pose}
\end{equation}
where $w_{diff}=e^{-\mu(S_{change})}$ adapts the weighting based on the quality difference score and $\mu$ is a decay parameter that controls the sensitivity of the weighting to scene changes. $T_{feature}$ represents the pose estimated by the \cite{c19} method, while $T_{direct}$ is the pose obtained through our direct method optimization. This adaptive weighting mechanism ensures that when the deviation from the reference benchmark is small, the system relies more on the feature-based pose; conversely, when the deviation is large, it gives greater weight to the direct method result. This approach provides smooth transitions between estimation methods and enhances system stability in challenging scenarios.
 
\section{EXPERIMENTAL RESULTS}

\begin{figure}[t]
  \centering
  \setlength{\fboxsep}{0pt} 
  \setlength{\fboxrule}{0pt} 
  \fbox{\includegraphics[width=0.95\linewidth]{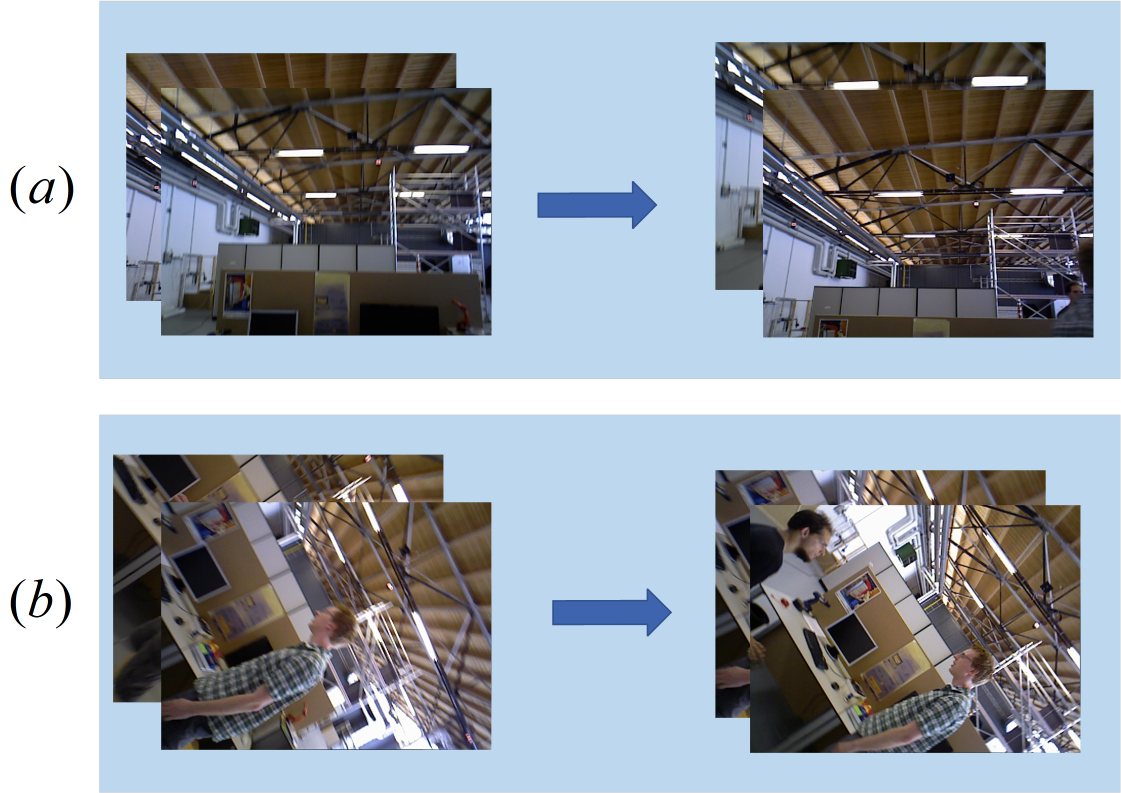}}
  \caption{\textbf{Scene quality assessment results on the $fr3/w/rpy$, showing detected problematic frames: (a) feature-poor ceiling, (b) rapid roll rotation.}}
  \label{fig:bad_scenes}
\end{figure}

\begin{figure*}[t]
  \centering
  \setlength{\fboxsep}{0pt} 
  \setlength{\fboxrule}{0pt} 
  \fbox{\includegraphics[width=0.85\linewidth]{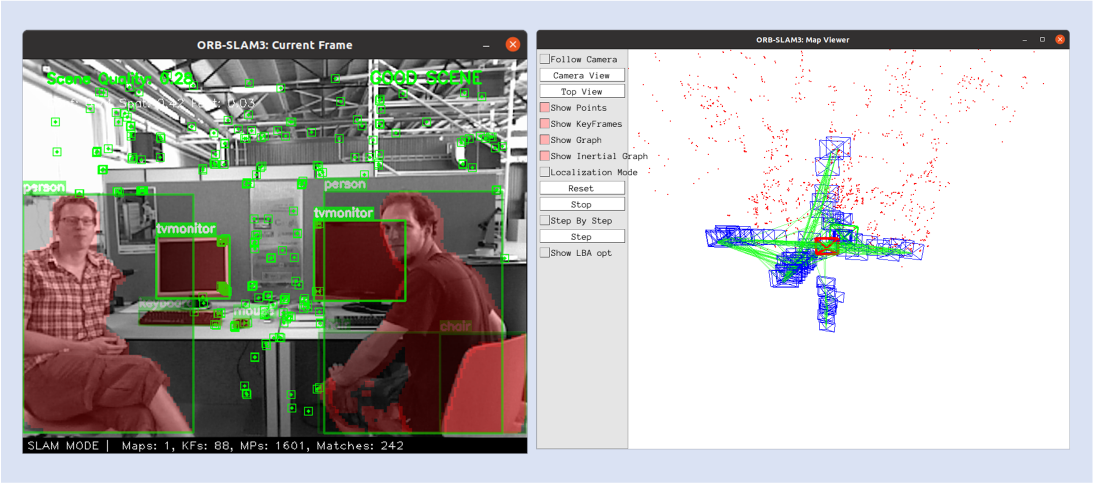}}
  \caption{\textbf{Visualization of our Scene-Object Quality Assessment integrated with ORB-SLAM3.}}
  \label{fig:SUQ_SLAM}
\end{figure*}

We evaluated our method on the TUM RGB-D datasets, widely used for benchmarking SLAM systems in dynamic environments.

\subsection{Evaluation of Scene-Object Quality Assessment Mechanism}

Our assessment mechanism identified 30 problematic frames out of 900 in the $fr3/w/rpy$ sequence. As Fig.~\ref{fig:bad_scenes} shows, these frames occurred when the camera moved toward the feature-poor ceiling (a) or during rapid roll rotations (b).

These results demonstrate the effectiveness of our assessment mechanism in identifying frames where traditional feature-based methods are prone to failure. The proposed quality metrics successfully capture geometric constraints (e.g., sparse features in ceiling views) and dynamic challenges (e.g., motion blur due to rapid rotation), validating the robustness of our approach.

\subsection{Comparison with State-of-the-Art Methods}

Fig.~\ref{fig:SUQ_SLAM} illustrates the integration of our system with ORB-SLAM3 on the TUM RGB-D dataset. The left panel presents the current frame with our Scene-Object Quality Assessment in action, where dynamic objects (e.g., people) are detected (green bounding boxes), segmented (red masks), and estimated using Lucas-Kanade optical flow to mitigate their impact on pose estimation. The scene quality score (0.36) and frame assessment (GOOD SCENE) are displayed in the upper corners, alongside individual quality metrics: confidence ($S_{conf}$), spatial distribution ($S_{spatial}$), and feature quality ($S_{feature}$) shown as white text. The right panel shows ORB-SLAM3’s sparse mapping results, including the 3D point cloud and camera pose. This visualization highlights how our framework enhances ORB-SLAM3 by improving scene quality assessment and handling dynamic objects, enabling more robust tracking in complex environments.

\begin{figure*}[t]
  \centering
  \setlength{\fboxsep}{0pt} 
  \setlength{\fboxrule}{0pt} 
  \fbox{\includegraphics[width=0.9\linewidth]{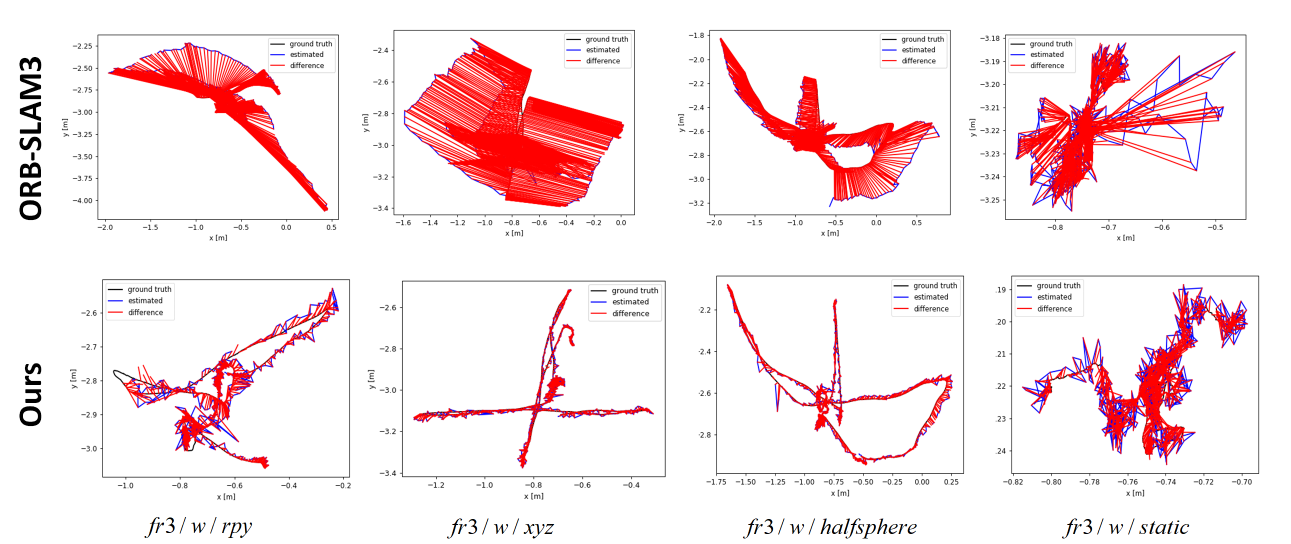}}
  \caption{\textbf{Visual comparison of estimated trajectories between ORB-SLAM3 and our method.}}
  \label{fig:Trajectory_compare}
\end{figure*}

We evaluated our system against several state-of-the-art SLAM methods for dynamic environments based on ORB-SLAM2 \cite{c20}, including DynaSLAM \cite{c13}, Blitz-SLAM \cite{c14}, and SG-SLAM \cite{c16}. Table~\ref{tab:combined_results_2} present the results for Absolute Trajectory Error (ATE) and Relative Pose Error (RPE) in translation and rotation. As shown in Table~\ref{tab:combined_results_2}, our method achieves competitive ATE performance and excels in frame-to-frame consistency and it consistently outperforms competing methods in translational and rotational RPE across most sequences. Notably, in challenging scenarios with complex camera motions, our system maintains high relative pose accuracy, demonstrating the effectiveness of our scene quality assessment and pose refinement strategies in dynamic environments.
\begin{table*}[!htbp]
\centering
\caption{COMPARATIVE RESULTS OF DIFFERENT SLAM SYSTEMS BASED ON ORB-SLAM2}
\label{tab:combined_results_2}
\scriptsize
\vspace{-2mm}
\setlength{\tabcolsep}{13pt}
\begin{tabular}{l|cc|cc|cc|cc|cc}
\hline
\rule{0pt}{2.5ex} 
& \multicolumn{10}{c}{\textbf{ABSOLUTE TRAJECTORY ERROR (ATE)}} \\
\cline{2-11}
\rule{0pt}{2.5ex}
Sequences & \multicolumn{2}{c|}{ORB-SLAM2} & \multicolumn{2}{c|}{Dyna-SLAM} & \multicolumn{2}{c|}{Blitz-SLAM} & \multicolumn{2}{c|}{SG-SLAM} & \multicolumn{2}{c}{Ours} \\
& RMSE & S.D. & RMSE & S.D. & RMSE & S.D. & RMSE & S.D. & RMSE & S.D. \\
\hline
\rule{0pt}{2.5ex}
\hspace{-1mm}fr3/w/half & 0.4543 & 0.2524 & 0.0296 & 0.0157 & \textbf{0.0256} & \textbf{0.0126} & 0.0268 & 0.0134 & 0.0286 & 0.0150 \\
fr3/w/xyz & 0.7521 & 0.4712 & 0.0164 & 0.0086 & 0.0153 & 0.0078 & \textbf{0.0152} & \textbf{0.0075} & \textbf{0.0152} & 0.0077 \\
fr3/w/rpy & 0.5391 & 0.2283 & 0.0354 & 0.0190 & 0.0356 & 0.0220 & 0.0324 & 0.0187 & \textbf{0.0302} & \textbf{0.0164} \\
fr3/w/static & 0.3194 & 0.1819 & 0.0068 & 0.0032 & 0.0102 & 0.0052 & \textbf{0.0073} & \textbf{0.0034} & 0.0087 & 0.0038 \\
fr3/s/static & - & - & - & - & - & - & \textbf{0.0060} & \textbf{0.0029} & 0.0067 & 0.0033 \\
fr3/s/xyz & \textbf{0.0092} & \textbf{0.0044} & 0.0127 & 0.0060 & 0.0148 & 0.0069 & - & - & 0.0172 & 0.0091 \\
\hline
\rule{0pt}{2.5ex} 
& \multicolumn{10}{c}{\textbf{METRIC TRANSLATIONAL DRIFT (T.RPE)}} \\
\cline{2-11}
\rule{0pt}{2.5ex}
\hspace{-1mm}fr3/w/half & 0.3216 & 0.2629 & 0.0284 & 0.0149 & 0.0253 & 0.0123 & 0.0279 & 0.0146 & \textbf{0.0159} & \textbf{0.0102} \\
fr3/w/xyz & 0.4834 & 0.3663 & 0.0217 & 0.0119 & 0.0197 & 0.0096 & 0.0194 & 0.0100 & \textbf{0.0125} & \textbf{0.0073} \\
fr3/w/rpy & 0.3880 & 0.2823 & 0.0448 & 0.0262 & 0.0473 & 0.0283 & 0.0450 & 0.0262 & \textbf{0.0252} & \textbf{0.0196} \\
fr3/w/static & 0.1928 & 0.1773 & 0.0089 & 0.0044 & 0.0129 & 0.0069 & 0.0100 & 0.0051 & \textbf{0.0063} & \textbf{0.0034} \\
fr3/s/static & - & - & - & - & - & - & 0.0075 & 0.0035 & \textbf{0.0058} & \textbf{0.0031} \\
fr3/s/xyz & \textbf{0.0117} & \textbf{0.0057} & 0.0142 & 0.0073 & 0.0144 & 0.0071 & - & - & 0.0124 & 0.0067 \\
\hline
\rule{0pt}{2.5ex} 
& \multicolumn{10}{c}{\textbf{METRIC ROTATIONAL DRIFT (ROT.RPE)}} \\
\cline{2-11}
\rule{0pt}{2.5ex}
\hspace{-1mm}fr3/w/half & 6.6515 & 5.3990 & 0.7842 & 0.4012 & 0.7879 & 0.3751 & 0.8119 & 0.3878 & \textbf{0.4324} & \textbf{0.2269} \\
fr3/w/xyz & 8.8419 & 6.6762 & 0.6284 & 0.3848 & 0.6132 & 0.3348 & 0.5040 & \textbf{0.2469} & \textbf{0.3878} & 0.2720 \\
fr3/w/rpy & 7.5906 & 5.4768 & 0.9864 & 0.5701 & 1.0841 & 0.6668 & 0.9565 & 0.5487 & \textbf{0.5868} & \textbf{0.4550} \\
fr3/w/static & 3.5991 & 3.2457 & 0.2612 & 0.1259 & 0.3038 & 0.1437 & 0.2678 & 0.1144 & \textbf{0.1770} & \textbf{0.0888} \\
fr3/s/static & - & - & - & - & - & - & 0.2657 & 0.1163 & \textbf{0.1580} & \textbf{0.0827} \\
fr3/s/xyz & 0.4874 & 0.2532 & 0.5042 & 0.2651 & 0.5024 & 0.2634 & - & - & \textbf{0.3392} & \textbf{0.1746} \\
\hline
\end{tabular}
\vspace{0.2mm}
\begin{flushleft}
\scriptsize \textbf{Note:} The best results of RMSE and S.D. are highlighted in bold.
\end{flushleft}
\end{table*}

We furtuer evaluated our system against DN-SLAM \cite{c21} based on ORB-SLAM3 \cite{c19}, as shown in Table~\ref{tab:combined_results_3}. Our method achieves a 92\%-97\% reduction in ATE over ORB-SLAM3 across dynamic sequences and performs comparably to DN-SLAM in absolute trajectory accuracy, with superior relative pose performance, particularly in the challenging $fr3/w/rpy$ sequence (RMSE: 0.025 vs. 0.065). These results demonstrate the effectiveness of our quality assessment and pose refinement strategies.
\begin{table*}[!htbp]
\centering
\caption{COMPARISON OF ATE AND T.RPE AMONG DIFFERENT SLAM SYSTEMS BASED ON ORB-SLAM3}
\label{tab:combined_results_3}
\scriptsize  
\vspace{-2mm}
\setlength{\tabcolsep}{7pt} 
\begin{tabular}{l|ccc|cc|ccc|cc}
\hline
\rule{0pt}{2.5ex}
& \multicolumn{5}{c|}{\textbf{ABSOLUTE TRAJECTORY ERROR (ATE)}} & \multicolumn{5}{c}{\textbf{METRIC TRANSLATIONAL DRIFT (T.RPE)}} \\
\cline{2-11}
\rule{0pt}{2.5ex} 
Sequences & ORB-SLAM3 & DN-SLAM & Ours & \multicolumn{2}{c|}{Improvements} & ORB-SLAM3 & DN-SLAM & Ours & \multicolumn{2}{c}{Improvements} \\
& RMSE (S.D.) & RMSE (S.D.) & RMSE (S.D.) & RMSE(\%) & S.D(\%) & RMSE (S.D.) & RMSE (S.D.) & RMSE (S.D.) & RMSE(\%) & S.D(\%) \\
\hline
\rule{0pt}{2.5ex} 
\hspace{-1mm}fr3/w/half & 0.362 (0.107) & \textbf{0.026} (\textbf{0.013}) & 0.028 (0.015) & 92.2\% & 85.9\% & 0.241 (0.177) & 0.035 (0.013) & \textbf{0.015} (\textbf{0.010}) & 93.7\% & 94.3\% \\
fr3/w/xyz & 0.454 (0.242) & \textbf{0.015} (\textbf{0.007}) & \textbf{0.015} (\textbf{0.007}) & 96.7\% & 97.1\% & 0.366 (0.276) & 0.024 (0.008) & \textbf{0.012} (\textbf{0.007}) & 96.7\% & 97.4\% \\
fr3/w/rpy & 0.761 (0.430) & 0.032 (0.019) & \textbf{0.030} (\textbf{0.016}) & 96.0\% & 96.2\% & 0.424 (0.240) & 0.065 (0.032) & \textbf{0.025} (\textbf{0.019}) & 94.1\% & 92.0\% \\
fr3/w/static & 0.027 (0.018) & \textbf{0.008} (\textbf{0.003}) & \textbf{0.008} (\textbf{0.003}) & 95.8\% & 85.6\% & 0.049 (0.023) & 0.011 (\textbf{0.003}) & \textbf{0.006} (\textbf{0.003}) & 87.8\% & 87.0\% \\
fr3/s/xyz & \textbf{0.009} (\textbf{0.005}) & 0.014 (0.008) & 0.017 (0.009) & -88.8\% & -80.0\% & \textbf{0.012} (\textbf{0.004}) & 0.015 (0.005) & \textbf{0.012} (0.006) & 0.0\% & -50.0\% \\
\hline
\end{tabular}
\vspace{0.2mm}
\begin{flushleft}
\scriptsize \textbf{Note:} The best results of RMSE and S.D. are highlighted in bold.
\end{flushleft}
\end{table*}
Fig.~\ref{fig:Trajectory_compare} compares estimated trajectories between ORB-SLAM3 (top) and our method (bottom) across four TUM sequences with three line tyrpes:  ground truth (black lines), estimated (blue lines), and difference (red lines).
Our method shows superior tracking accuracy, with reduced error in all sequences. It maintains near-perfect alignment with ground truth in $fr3/w/xyz$ and accurately tracks complex movements in $fr3/w/halfsphere$ and $fr3/w/rpy$ under dynamic conditions.

Our experiments confirm that our scene quality assessment and pose refinement strategies enhance SLAM performance in dynamic environments. While achieving competitive absolute trajectory accuracy, our method excels in frame-to-frame consistency, as demonstrated by superior RPE metrics across most sequences.

\section{CONCLUSIONS}

In this paper, we presented an adaptive prior scene-object SLAM framework for dynamic environments. Our quality assessment mechanism effectively identifies problematic frames, while our direct pose refinement strategy corrects tracking errors when traditional methods fail. Experimental results show significant improvements over state-of-the-art methods, particularly in maintaining consistent tracking.

For future work, we plan to enhance the system by incorporating line and plane features \cite{c22} to improve quality assessment, especially in texture-poor environments. We also aim to develop a global temporal management strategy for good and bad frames based on data association \cite{c23}, enabling comprehensive optimization across the entire trajectory rather than just frame-to-frame refinement. These advancements will further enhance system robustness in complex dynamic scenarios and support longer-term consistent mapping.

\end{document}